\documentclass{article} % For LaTeX2e
\usepackage{iclr2024_conference,times,amsmath}

\usepackage{amsmath,amsfonts,bm}

\def\eqref#1{equation~\ref{#1}}
\def\1{\bm{1}}

\DeclareMathAlphabet{\mathsfit}{\encodingdefault}{\sfdefault}{m}{sl}
\SetMathAlphabet{\mathsfit}{bold}{\encodingdefault}{\sfdefault}{bx}{n}

\DeclareMathOperator*{\argmin}{arg\,min}

\usepackage{hyperref}
\usepackage{url}
\usepackage{graphicx}
\usepackage{booktabs}

\title{Enhancing Joint Motion Prediction for Individuals with Limb Loss Through Model Reprogramming}

\author{Sharmita Dey \thanks{ Corresponding author \\
Sharmita Dey conceptualized and led the project. Both authors contributed equally to the methodology, design, and writing of the manuscript.\\
Time Series Representation Learning for Health @ ICLR }\\
Department of Computer Science\\
University of Goettingen, Germany \\
\texttt{sharmita.dey@cs.uni-goettingen.de} \\
\And
Sarath R. Nair \\
University Medical Center Goettingen \\
Germany \\}

\iclrfinalcopy % Uncomment for camera-ready version, but NOT for submission.
\begin{document}

\maketitle

\begin{abstract}
Mobility impairment caused by limb loss is a significant challenge faced by millions of individuals worldwide. The development of advanced assistive technologies, such as prosthetic devices, has the potential to greatly improve the quality of life for amputee patients. A critical component in the design of such technologies is the accurate prediction of reference joint motion for the missing limb. However, this task is hindered by the scarcity of joint motion data available for amputee patients, in contrast to the substantial quantity of data from able-bodied subjects.  To overcome this, we leverage deep learning's reprogramming property to repurpose well-trained models for a new goal without altering the model parameters. With only data-level manipulation, we adapt models originally designed for able-bodied people to forecast joint motion in amputees. The findings in this study have significant implications for advancing assistive tech and amputee mobility.
\end{abstract}

\section{Introduction}
Limb loss is a profound challenge affecting millions globally, significantly hindering mobility and daily activities. This condition introduces physical and psychological limitations, impacting individuals' well-being and independence. In response, substantial efforts have been directed towards developing assistive technologies, particularly advanced prosthetics, to improve amputees' lives by mimicking natural limb functions \citep{gehlhar2023review}.
Accurate joint motion prediction for the missing limb is critical in designing effective assistive devices \citep{windrich2016active, dey2019random, dey2020continuous}. However, this task faces obstacles due to the limited data available for amputee patients and the diverse nature of amputations, which vary greatly among individuals \citep{dey2020feasibility}. Traditional data-driven modeling approaches used for lower-limb joint motion prediction struggle to address these issues due to data scarcity and the need for models to account for the wide range of motion patterns and functional variations resulting from different amputation types.

In this context, "model reprogramming" emerges as a transformative technique that holds promise in overcoming the challenges posed by data scarcity for amputee patients. Model reprogramming is an approach that repurposes a machine learning model originally trained for one domain to perform a task in a different chosen domain, without necessitating extensive retraining or fine-tuning \citep{elsayed2018adversarial}. It presents a cost-effective approach, requiring fewer parameter adjustments compared to developing new subject-specific models, thus reducing computational demands while preserving the model's original capabilities. The efficiency of this reprogramming approach has been validated in the context of image classification \citep{elsayed2018adversarial,tsai2020transfer} as well as time-series analysis \citep{hambardzumyan2021warp,huck2021voice2series} and out-of-distribution detection \citep{wang2022watermarking}. This research explores the potential of model reprogramming to address the challenges of predicting joint motion in amputees, with the aim of enhancing the development of assistive technologies. By repurposing well-established gait prediction models from able-bodied data, we seek to provide amputees with better mobility solutions, potentially transforming the landscape of prosthetic development and offering a higher quality of life for individuals facing limb loss.

\section{Method}
\label{gen_inst}

\textbf{Datasets}. 
The research utilized a public dataset \citep{hu2018benchmark} for training the gait prediction model on able-bodied individuals, employing time series kinematic sequences from Inertial Measurement Units (IMUs) and goniometers to capture 3D movements and limb positions. The dataset encompassed ten subjects performing various activities, resulting in over 5 million samples for training and 1.3 million for testing. Additionally, data was acquired and processed from three transtibial amputees in a gait laboratory using a Vicon Motion Systems setup with retro-reflective markers for 3D tracking across different locomotion tasks. Gait cycle boundaries were determined using Vicon Nexus software, and joint angles and kinematic data were computed with OpenSim \citep{delp2007opensim}. Informed consent was obtained from all participants.

\textbf{Approach}. We aim to predict the walking patterns of amputees' missing limbs using a model trained on able-bodied motion data, without modifying the model's parameters. By reprogramming amputee input data into compatible time series sequences, we adapt them for the model, enabling accurate motion predictions. Our modeling includes three key components: a \textit{foundation module}, a versatile model trained on a diverse able-bodied dataset; a \textit{template mapping} algorithm for adjusting amputee inputs based on a correction template; and a \textit{refurbish module} that aligns amputee inputs with this template. The foundation module uses the adjusted inputs from amputees to predict the motion of the missing limb. This architecture is designed to seamlessly integrate amputee data into the model, ensuring accurate limb motion prediction. Fig.  \ref{fig:template_mapping} summarizes this architecture.

\subsection{Foundation module}
We developed a multi-task foundation model trained on data from able-bodied individuals, capturing various motion scenarios. To exploit commonalities across tasks while catering to each task's unique needs, we design a model $g(.)$ which includes a shared core $g_s(.)$ with parameters $\theta_s$ shared across tasks, and task-specific layers, $g_t(.)$ with parameters $\theta_t$ tailored to each individual task. The model's prediction for an input $X_{t, k}$ belonging to task $t$ at timepoint $k$ is $g(X_{t, k}, t; \theta_g = \{\theta_{s}, \bigcup_{t}\theta_t\}) = g_t(g_s(X_{t, k}; \theta_s); \theta_{t})$. The shared core $g_s$ employs a time convolution neural network (TCN, \citep{bai2018empirical}) for extracting the temporal features, while the task-specific layers $g_t$ use a two-layer MLP with ReLU activation to fine-tune the output for each specific task's requirements.

\subsection{Template mapping}

\begin{figure*}
    \centering
    \includegraphics[width=0.9\textwidth]{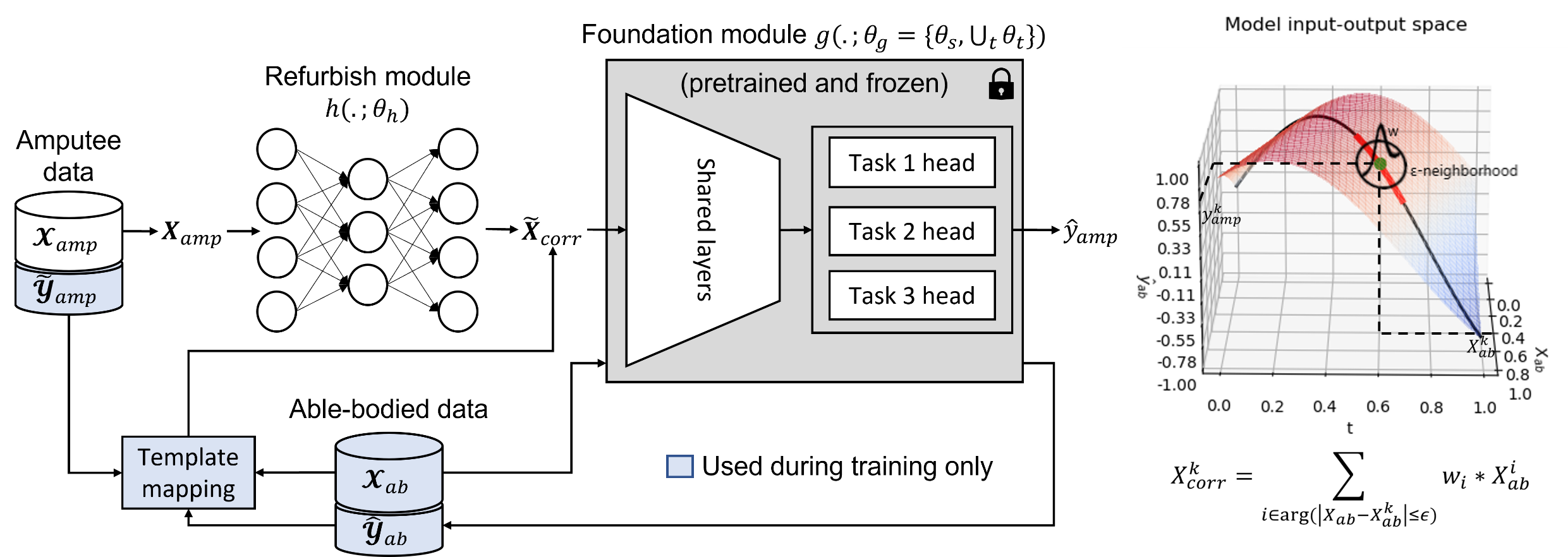}
    \caption{Illustration of computation of the correction input $X_{corr}$ corresponding to the $k$-th input sample $X^k_{amp}$ of the amputee. The able-bodied input $X^i_{ab}$ that produces the most similar output as that of the desired amputee output $y_{amp}^k$ is searched in the input-output space of the trained able-bodied foundation module. Instead of searching based on a single desired motion variable $y_{amp}^k$, a sequence of values $\{y^{k-m}_{amp}, ..., y^k_{amp}, ..., y^{k+m}_{amp}\}$ (marked by the red region in the right) is used and the able-bodied input $X^i_{ab}$ corresponding to the midpoint of the sequence is considered. Further, a neighborhood of radius $\epsilon$ is considered around $X^i_{ab}$ and the correction input $X_{corr}$ is computed as a weighted sum of samples in this neighborhood with weights decreasing (linearly or exponentially) with increasing distance from the center $X^i_{ab}$. }
    \label{fig:template_mapping}
\end{figure*}

Once the able-bodied model is trained, we use it to predict the motion variables for amputee subjects without retraining the model with amputee data. To achieve this, we map the amputee inputs $X_{amp}$ corresponding to a desired output $y_{amp}$ to a corrected input $X_{corr}$ such that the model gives a similar output as the amputee's desired output for this corrected input (Fig. \ref{fig:template_mapping}). Note that the desired amputee output ${y}^k_{amp}$ is computed from those able-bodied individuals sharing similar anthropometry (such as height, mass, and age) as that of the amputee. 
A naive way of mapping the amputee input to an able-bodied template is to use the able-bodied input $X^k_{ab}$ for which the model gives the desired amputee output ${y}^k_{amp}$ as the corrected input $X^k_{corr}$. 
This mapping approach faces temporal ambiguity when identical desired outputs $y^k_{amp}$ appear at different times, potentially mapping diverse able-bodied inputs to the same amputee input values (for example, $y^k_{amp}=0.6$ at $t=0.2$ and $t=0.6$ leading to different $X^k_{corr}$ for same $y^k_{amp}$). 
To address temporal ambiguity, we consider a sequence of co-occurring values in a time of which the desired output $y^k_{amp}$ is the midpoint (red region in Fig. \ref{fig:template_mapping}). 
The correction input $X^k_{corr}$ in this case can be computed as 

\begin{equation}
    X^k_{corr} = \argmin_{X^i_{ab}} \sum_{j=-m}^m \lVert f(X^{i-j}_{ab}) - y^{k-j}_{amp} \rVert
    \label{eq:sequence_mapping}
\end{equation}

In the above, we use a single able-bodied input for amputee correction, which may lead to noise and overfitting. To deal with this, we propose using multiple able-bodied inputs within a defined $\epsilon$-neighborhood around the closest match (computed in Eq. \ref{eq:sequence_mapping}) to the desired amputee output (black ellipse in Fig. \ref{fig:template_mapping} right). 
For simplicity, we consider only $n$ closest neighbors within the $\epsilon$-neighborhood for computing the correction template $X_{corr}$. 

\subsection{Refurbish module} 
We adapt amputee inputs $X_{amp}$ to correct for the compensatory motions and asymmetric gait by learning a mapping to a correction template $X_{corr}$, derived from able-bodied data. This is done using a refurbish module $h(.)$ to train on the relationship between $X_{amp}$ and $X_{corr}$ by minimizing two loss functions: 

\begin{equation}
\begin{aligned}
    \mathcal{L}(\Theta_{h}) &= \alpha\sum_{X_{amp}, X_{corr} \in \mathcal{X}^{train}_{amp}} \lVert h(X_{amp}; \Theta_{h}) - X_{corr} \rVert \\&+ \beta\sum_{X_{amp}, y_{amp} \in \mathcal{X}^{train}_{amp}} \lVert g(h(X_{amp}; \Theta_{h}), \Theta^*_g) - y_{amp} \rVert
\label{eq: correction_based_loss}
\end{aligned}
\end{equation}
where the first term on the R.H.S is the (mean squared error) MSE loss between the computed correction $X_{corr}$ (from the template mapping step) and refurbish module predicted correction $\hat{X}_{corr}=h(X_{amp}; \Theta_{h})$, the second term on the R.H.S is the MSE loss between the predictions of the pre-trained frozen foundation module $g(., \Theta^*_g)$ when it is fed with the refurbish module's outputs and the desired target values of the amputee $y_{amp}$, $\alpha$ and $\beta$ are the factors that influence the effect of the two terms on final loss. %In this strategy, the computed correction templates are completely ignored, and the model is trained solely to reproduce the final desired target. The loss function becomes

\section{Experiments}

\textbf{Baselines. }
We assess our proposed method's effectiveness against two baseline approaches: (1) \textit{cross-mapping}, where, we directly apply the foundation model $g(.)$ to predict motion variables for amputee subjects, without utilizing the refurbish module to transform amputee inputs into clean data. (2) \textit{direct-mapping}, where we train user-specific models to directly learn the mapping from amputee inputs $X_{amp}$ to the desired motion variables $y_{amp}$.

\begin{figure}
    \centering
    \includegraphics[width=0.7\linewidth]{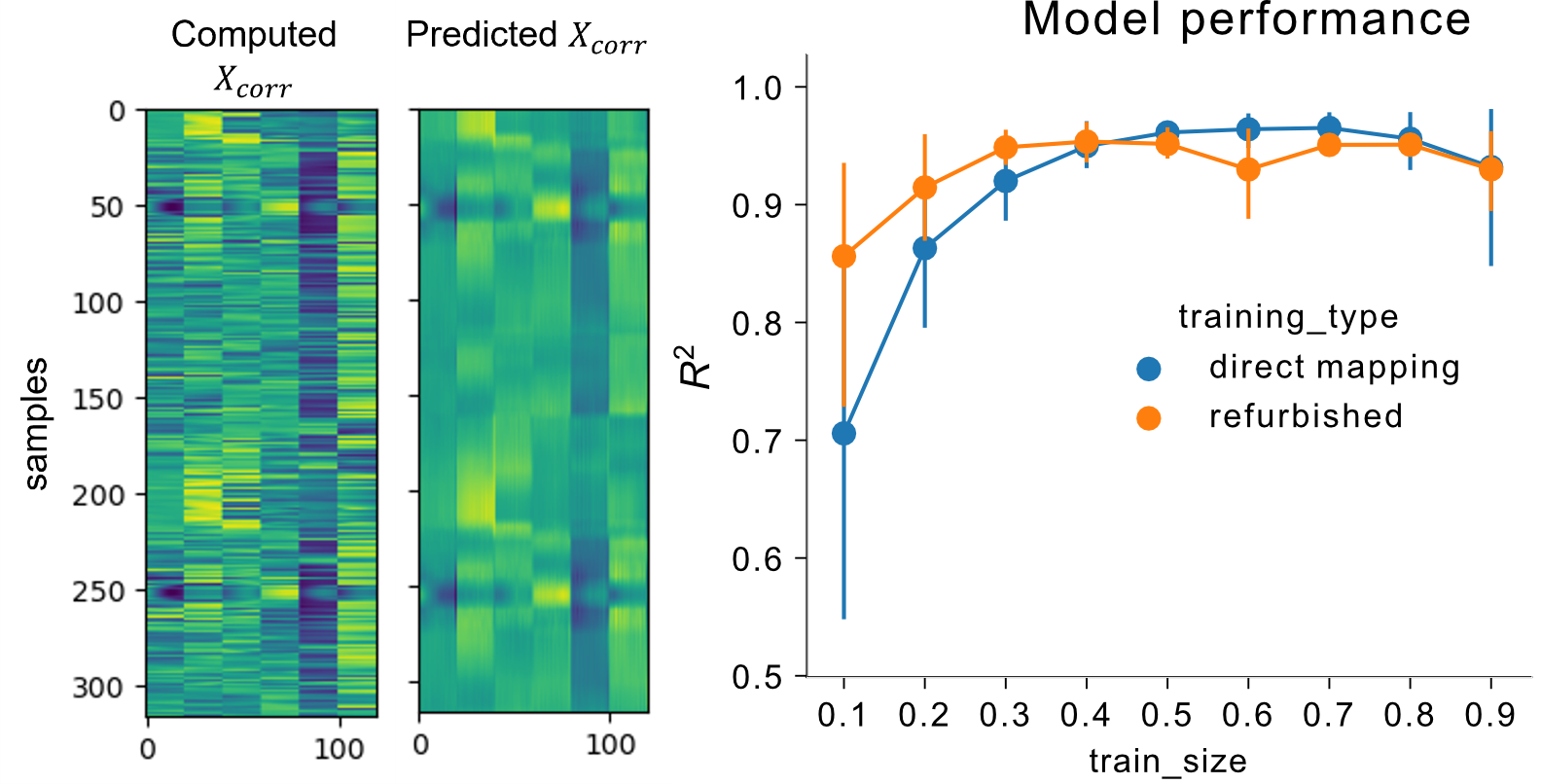}
    \caption{(Left) The computed correction template $X_{corr}$ and corresponding prediction by the refurbish module. (Right) Performance of models trained with direct mapping and refurbishing. For refurbishing, $2m+1=1$, $n=5$ with linear weighting, $\alpha=1$ and $\beta$ = 20 were selected based on an extensive grid search. where $2m+1$ is the length of the sequence considered for template matching and $n$ is the number of closest neighbors within the $\epsilon$-neighborhood used for computing the correction template. }
    \label{fig:results}
\end{figure}

\textbf{Results.} 
We compared the performance of a model that is fed with refurbished inputs to the direct mapping method for different amounts of training data (Fig. \ref{fig:results} right). A general improvement in prediction scores was observed for both methods as the training sample size increased. There was a slight drop in accuracy when using very large training sample sizes, likely due to overfitting. Most strikingly, the proposed refurbishing approach outperformed direct mapping for smaller amounts of training data (direct mapping reached similar accuracies as that of refurbishing only when 40\% of total amputee data was used for training). 
For the lowest amount of training data, refurbishing provided an improvement in performance of around 15\% compared to direct mapping. On the other hand, cross-mapping (zero-shot inference on amputee data using models trained on able-bodied subjects) performed poorly (Tab. \ref{tab:results_summary}). Additionally, we found that the output of the trained refurbish module (that is fed to the foundation module for target prediction) matched closely with the correction input $X_{corr}$ computed by the template mapping procedure aiding the model to produce accurate predictions (Fig. \ref{fig:results} left). 
These results suggest that our proposed input refurbishing strategy offers an efficient means to adapt a pre-trained model for new scenarios, especially when dealing with limited training data.

% \usepackage{booktabs}
% \usepackage{graphicx}
\begin{table}[!htbp]
\caption{$R^2$ scores obtained with different training strategies for a train sample ratio of 0.1}
\label{tab:results_summary}
\centering
\resizebox{0.5\columnwidth}{!}{%
\begin{tabular}{@{}ccc@{}}
\toprule
 \begin{tabular}[c]{@{}c@{}}cross\\ mapping\end{tabular} & \begin{tabular}[c]{@{}c@{}}direct \\ mapping\end{tabular} & \begin{tabular}[c]{@{}c@{}}refurbished \\ ($\alpha=1, \beta=20$)\end{tabular} \\ \midrule
-0.32 $\pm$ 0.48 & 0.71 $\pm$ 0.20 & \textbf{0.86 $\pm$ 0.14} \\ \bottomrule
\end{tabular}%
}
\end{table}

While direct mapping performs well in predicting motion variables for the specific amputee it was trained on, it necessitates a larger dataset to achieve the same level of performance as the model reprogramming approach. Additionally, direct mapping has limitations as it relies on limited data from amputees, resulting in reduced exposure to diverse motion conditions. Consequently, it may exhibit limited generalization capabilities, a crucial quality for amputee motion prediction models, given the potential for changes in gait patterns as amputees adapt to generated motion. In contrast, an able-bodied model trained on a wide array of motion scenarios from various individuals may be more adaptable to evolving amputee gait patterns while accommodating the predicted joint motion required for prosthetic walking.

 \section{Conclusion}

 This study introduces a method to repurpose a generic, well-trained model for predicting amputees' gait variables without modifying the model's parameters. By incorporating a lightweight module for data manipulation, we utilize the existing model to accurately predict amputee-specific gait variables. Our results show that this input refurbishing technique efficiently adapts pre-trained models to new applications, particularly useful in scenarios with limited training data. This approach promises significant improvements in model performance and versatility, with potential applications in controlling powered prostheses.

  \section{Acknowledgements}
  The authors would like to thank Dr. Thomas Schmalz and Dr. Michael Ernst (Department of Clinical Research \& Services, Ottobock SE \& Co. KGaA, Germany) and Dr. Takashi Yoshida (University Medical Center Goettingen) for their assistance in conducting experiments with the transtibial amputees and collecting the data.

\bibliography{iclr2024_conference}
\bibliographystyle{iclr2024_conference}

\appendix
%\section{Appendix}
%You may include other additional sections here.

\end{document}